\documentclass[final]{cvpr}

\usepackage{times}
\usepackage{epsfig}
\usepackage{graphicx}
\usepackage{amsmath}
\usepackage{amssymb}
\usepackage[symbol]{footmisc}

\usepackage[pagebackref=true,breaklinks=true,colorlinks,bookmarks=false]{hyperref}

\def\cvprPaperID{8240} % *** Enter the CVPR Paper ID here
\def\confYear{CVPR 2021}

\begin{document}

\title{ParaNet: Deep Regular Representation for 3D Point Clouds}

\author{

% Qijian Zhang\quad Junhui Hou\footnotemark[1] \quad Yue Qian \\
% Department of Computer Science,
% City University of Hong Kong\\
% {\tt\small \{qijizhang3-c@my., jh.hou@, yueqian4-c@my.\}cityu.edu.hk}

% \and
% Juyong Zhang\\
% School of Mathematical Sciences\\
% University of Science and Technology of China\\
% {\tt\small juyong@ustc.edu.cn}

% \and
% Ying He\\
% School of Computer Science and Engineering\\
% Nanyang Technological University\\
% {\tt\small yhe@ntu.edu.sg}

% Qijian Zhang$^1$\quad Junhui Hou$^1$\footnote{Corresponding author.}\quad Yue Qian$^1$ \quad Juyong Zhang$^2$ \quad Ying He$^3$\\
% $^1$Department of Computer Science, City University of Hong Kong\\
% $^2$School of Mathematical Sciences,
% University of Science and Technology of China\\
% $^3$School of Computer Science and Engineering,
% Nanyang Technological University\\
% {\tt\small \{qijizhang3-c@my., jh.hou@, yueqian4-c@my.\}cityu.edu.hk, juyong@ustc.edu.cn,  yhe@ntu.edu.sg}

Qijian Zhang$^1$\quad Junhui Hou$^{1*}$\quad Yue Qian$^1$ \quad Juyong Zhang$^2$ \quad Ying He$^3$\\
$^1$Department of Computer Science, City University of Hong Kong\\
$^2$School of Mathematical Sciences,
University of Science and Technology of China\\
$^3$School of Computer Science and Engineering,
Nanyang Technological University\\
{\tt\small \{qijizhang3-c@my., jh.hou@, yueqian4-c@my.\}cityu.edu.hk, juyong@ustc.edu.cn,  yhe@ntu.edu.sg}

}
\maketitle
\footnotetext[1]{Corresponding author.}

%%%%%%%%% ABSTRACT
\begin{abstract} \label{abs}
	Although convolutional neural networks have achieved remarkable success in analyzing 2D images/videos, it is still non-trivial to apply the well-developed 2D techniques in regular domains to the irregular 3D point cloud data. To bridge this gap, we propose ParaNet, a novel end-to-end deep learning framework, for representing 3D point clouds in a completely regular and nearly lossless manner. To be specific, ParaNet converts an irregular 3D point cloud into a regular 2D color image, named point geometry image (PGI), where each pixel encodes the spatial coordinates of a point. In contrast to conventional regular representation modalities based on multi-view projection and voxelization, the proposed representation is differentiable and reversible. Technically, ParaNet is composed of a surface embedding module, which parameterizes 3D surface points onto a unit square, and a grid resampling module, which resamples the embedded 2D manifold over regular dense grids. Note that ParaNet is unsupervised, i.e., the training simply relies on reference-free geometry constraints. The PGIs can be seamlessly coupled with a task network established upon standard and mature techniques for 2D images/videos to realize a specific task for 3D point clouds. We evaluate ParaNet over shape classification and point cloud upsampling, in which our solutions perform favorably against the existing state-of-the-art methods. We believe such a paradigm will open up many possibilities to advance the progress of deep learning-based point cloud processing and understanding.
\end{abstract}

\section{Introduction} \label{introduction}

\begin{figure}[t]
	\centering
	\includegraphics[width=1\linewidth]{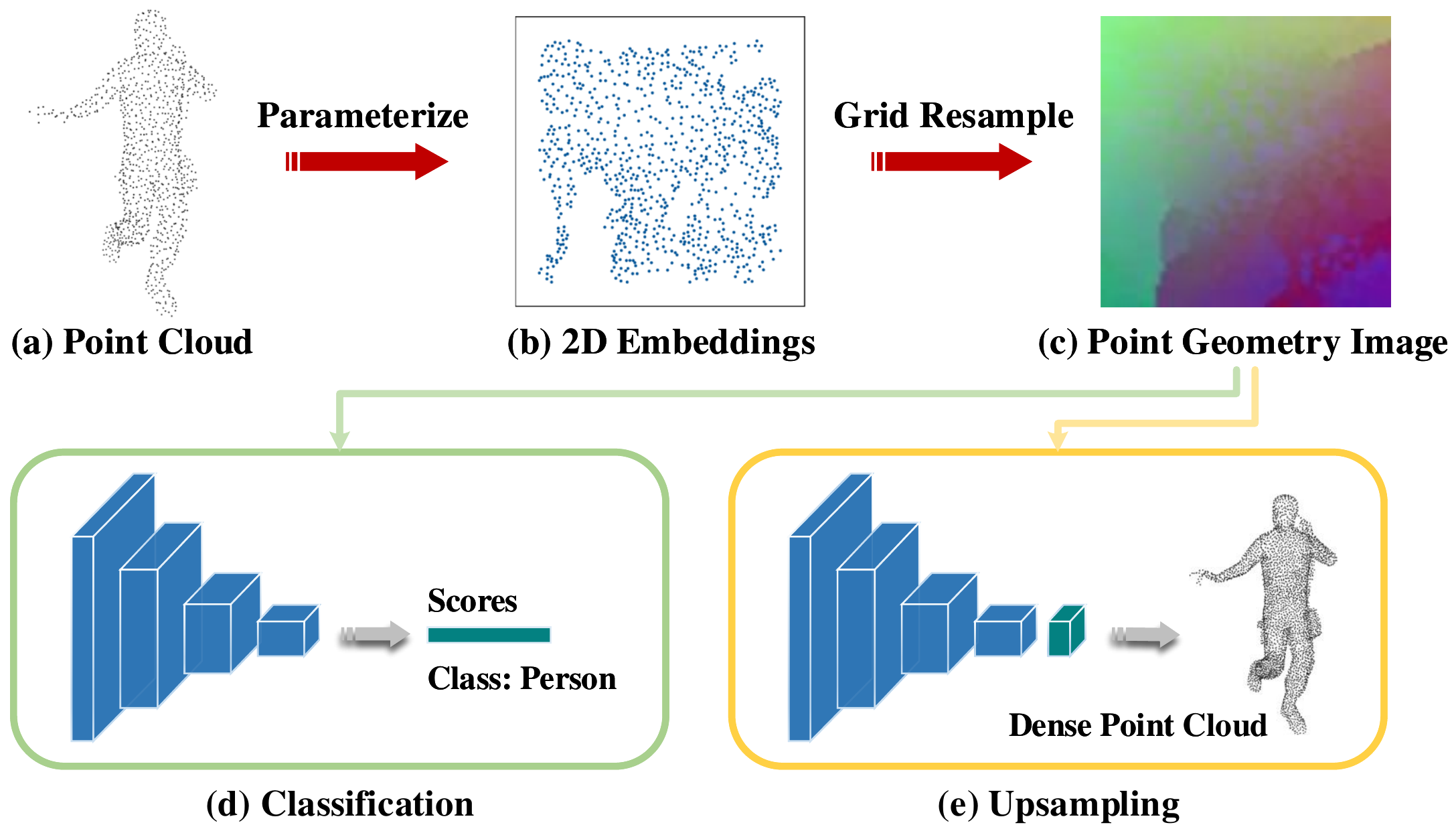}
	\caption{Illustration of the pipeline of ParaNet. ParaNet maps an input 3D point cloud (a) to a 2D domain (b) and adopts grid resampling to produce a 2D color image named point geometry image (PGI) (c), which is a complete regular structure that explicitly encodes 3D information into pixel colors and can faithfully represent the geometry of the input shape. Thanks to the regular structure, the generated geometry images enable the standard 2D CNNs for 3D applications, such as classification (d) and upsampling (e). Note that ParaNet is unsupervised, i.e., the ground-truth PGIs are not required during training.}
	\label{Figure-01}
\end{figure}

The recent decade has witnessed remarkable success of deep convolutional neural networks (CNNs) in analyzing regular visual modalities such as 2D images/videos and 3D volumes
\cite{RegCNN-1, RegCNN-2, RegCNN-3, RegCNN-4, RegCNN-5, RegCNN-6, RegCNN-7, RegCNN-8}.
In the meantime, with the advanced 3D sensing technology, 3D point clouds become a popular scheme for representing 3D objects and scenes and have been used in a wide range of applications, including immersive telepresence~\cite{orts2016holoportation}, 3D city
reconstruction~\cite{lafarge2012creating, musialski2013survey}, autonomous
driving~\cite{chen2017multi, li20173d}, and virtual/augmented reality~\cite{held20123d,santana2017multimodal}. However, due to the irregular and orderless nature of point clouds, applying the well-developed CNNs is highly non-trivial.

In order to apply the well-developed CNNs to irregular 3D point clouds, there are two main classes of approaches: rasterization-based methods and point-based methods.
The former converts the input model into either a regular 3D volume via voxelization or a collection of multi-view 2D images using projection. Though one can directly apply the standard 2D/3D CNNs to the regularly rasterized data, their performances are limited by either irreversible information loss or high computational complexity.
The latter takes 3D point clouds directly as input, and extracts features from each point. For example, the pioneering work, PointNet~\cite{PointNet}, 
can be viewed as $1\times 1$ convolution applied to each point.
Many follow-up works aim at designing convolution-like operators on point clouds
for feature extraction~\cite{SPLATNet,BiConv,SO-Net,FeaStNet,PointCNN,PointConv}. Each point-based convolution operator has its own merits and limitations. Currently, there is no unified framework adopted by the community.  Moreover, it is observed  that in point-based networks most of the computational resources are wasted on the process of neighborhood construction (e.g., $k$-NN) and structurization, not on the actual feature extraction.

\begin{figure}[t]
	\centering
	\includegraphics[width=0.95\linewidth]{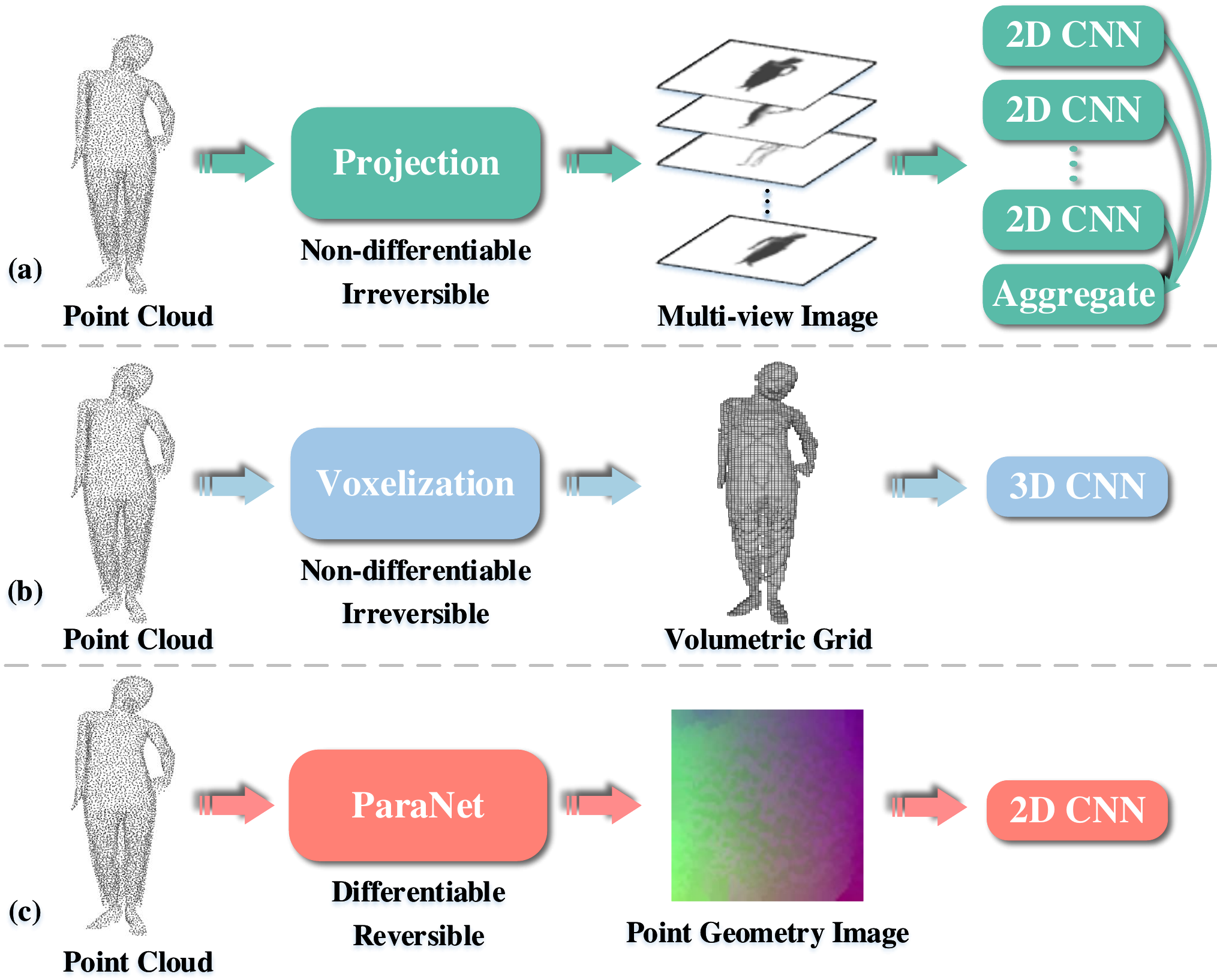}
	\caption{Illustration of three classes of methods for applying the regular-data-fed CNNs to 3D point clouds.  (a) Projection-based methods take 3D point clouds as a collection of multi-view images, and adopts multiple 2D CNNs with additional viewpoint aggregation. (b) Voxelization-based methods quantize point clouds as volumetric grids, which are processed by 3D CNNs. (c) ParaNet converts the input model into a PGI and adopts 2D CNNs. }
	\label{Figure-02}
\end{figure}

Instead of seeking for convolution-like operators on unstructured points, we consider a different paradigm that converts an irregular 3D point cloud to a completely regular 2D representation where % so that
standard 2D CNNs can be applied. In the computer graphics community, there are extensive research on mesh parameterization. When mapping a 3D surface to a 2D domain, distortion is inevitable unless the surface is developable. Therefore, most of the existing techniques formulate parameterization as an optimization problem, aiming at finding a balance between distortion and parameterization quality (e.g., the number of singularities, continuity along cuts, etc). Although the state-of-the-art methods~\cite{DBLP:journals/tog/CampenBK15,DBLP:journals/tog/LyonCBK19} can compute globally smooth and seamless parameterization with low angle and area distortions, these approaches highly rely on the explicit connectivity information encoded in meshes, extending them to point clouds is highly non-trivial. Moreover, they usually work for shapes with simple geometry and topology, since the complexity and difficulty in solving the optimization problem increases significantly with the complexity of geometry and topology.

In this paper we propose an end-to-end deep learning-based framework, called ParaNet, which is capable of transforming an arbitrary 3D point cloud into a completely regular 2D structure dubbed as point geometry image (PGI), which encodes the Cartesian coordinates of 3D points $(x, y, z)$ as the color values of pixels $(r, g, b)$. See Fig. \ref{Figure-01}. 
Since the generated PGIs can faithfully represent the input shape and naturally fit into the existing CNN pipelines, it bridges the gap between 3D geometry and the powerful deep learning techniques well-investigated for regular data structures.
Technically, we design ParaNet as a two-stage network and train it in an end-to-end unsupervised fashion. In the first stage, the points of an input shape pass through a surface embedding module and are parameterized into a unit square while maintaining local proximity as much as possible.  
In the second stage, the grid resampling module converts the 2D samples to a PGI. 
The ParaNet can be seamlessly coupled with downstream task networks established upon standard convolution techniques to boost various point cloud applications such as classification and upsampling.

\textbf{Summary of Contributions}. ParaNet is the first end-to-end deep neural network for representing an irregular 3D point cloud as a regular 2D image amenable to conventional CNNs driven by the standard convolution. Compared with multi-view projection and voxelization based approaches~\cite{MVCNN, PCN, Volumetric_and_Multiview,ModelNet, VoxNet,OctNet, OCNN,KD-Net}, our PGI-based representation modality is more accurate and maintains reversible, which means that the original point clouds can be directly restored without borrowing additional computations. Furthermore, as illustrated in Fig.~\ref{Figure-02}, the generation process of PGI is differentiable and can be optimized by a deep learning framework, whereas rasterization-based methods adopt non-differentiable quantization. To validate its potential in point cloud analysis, we combine ParaNet with two downstream task networks, classification and patch-based upsampling, and obtain favorable performance compared with state-of-the-art methods. The proposed regular point cloud representation modality enables  effortless adaptation of the mature visual processing techniques originally developed for 2D images/videos, making the wide range of 2D computational tools available to the point cloud community.

\section{Related Work} \label{rw}

\subsection{Rasterization-based Point Cloud Analysis}
\label{sec_2_1}
View projection \cite{MVCNN, PCN, Volumetric_and_Multiview} generates a set of 
% rendered views on 
2D images by setting virtual cameras around 3D shape. Although multi-view projection is able to utilize well-studied 2D CNN for feature extraction, it suffers from information loss~\cite{KD-Net} and is not easily extended for fine-grained 3D tasks like segmentation~\cite{ScanNet}. Another rasterization-based scheme is to voxelize point clouds into quantized volumetric grids followed by standard 3D CNNs \cite{ModelNet, VoxNet,OctNet, OCNN,KD-Net}. However, voxelization suffers from inaccurate data representation, or heavy computation and large memory expense, depending on the resolution.

\subsection{Convolution on Point Clouds}
To eliminate quantization loss, latest researches focus on designing deep learning based architectures directly operated on point clouds without any rasterization procedure as pre-processing. Qi \textit{et al.} \cite{PointNet} developed the pioneering point-based deep set architecture, \textit{i.e.}, PointNet, that employs shared multi-layer perceptrons for point-wise feature embedding and achieves permutation-invariance by max-pooling, which is improved in PointNet++ \cite{PointNet++} by incorporating neighbor information to model local and global geometry hierarchically. Based on such point-wise feature learning, more methods~\cite{SPLATNet,BiConv,SO-Net,FeaStNet,PointCNN,PointConv} were proposed with different design in local feature aggregations and hierarchical learning strategies.
In general, existing studies devote to tailoring \textit{convolution-like} operators and \textit{CNN-like} architectures for point cloud feature extraction. Compared with rasterization-based methods, point-based methods does not have quantization issue and can achieve better performance. Unlike image analysis in which 2D CNNs are dominating, currently there is no standard framework for 3D point clouds. Moreover, directly processing irregular points is computationally expensive, due to extra resources for structurization and neighborhood construction~\cite{PVCNN}.

\subsection{Parameterization and Geometry Images} \label{sec_2_3}
In digital geometry processing, parameterization is a typical way to convert an irregular representation (e.g., polygonal mesh, point clouds) to a regular one. Since a map from 3D surfaces to 2D domains inevitably has distortion unless the surface is developable, conformal parameterization is popular thanks to its angle preserving property~\cite{DBLP:conf/sgp/GuY03}.
It is well known that area distortion of a conformal parameterization is closely related to the singularities of parameterization, i.e., the number, their locations and indices. 
Modern approaches often formulate conformal parameterization as a mixed-integer programming~\cite{DBLP:journals/tog/BommesZK09}, in which the locations of singularities are continuous, and their numbers and indices are discrete. Once the parameterization is computed, one can generate a multi-chart geometry image by partitioning the model via a motorcycle graph~\cite{DBLP:journals/tog/CampenBK15}. These methods are theoretically sound and elegant, however, they are not practical for surfaces with complex geometry and/or topology due to high computational cost. Moreover, these approaches work only for single-component manifold surfaces, whereas many man-made models are non-manifolds with multiple components. Note that the proposed point geometry images are fundamentally different from the classic geometry images in terms of generation and application.

\if 0
\rc{
	Sinha \textit{et al.} \cite{DL-GI} validate the viability of applying CNNs to learn shape abstractions from geometry images, generated by an improved remeshing approach. Differently, our solution directly operates on point clouds without meshes while maintaining simplicity and efficiency.
}
\fi

\subsection{Deformation-based 3D Shape Representation} \label{sec_2_4}

FoldingNet, proposed by Yang \textit{et al.} \cite{FoldingNet}, is a novel deep auto-encoder. The folding-based decoder, which consists of shared multi-layer perceptrons, deforms a predefined 2D regular grid to the input 3D point cloud.
The idea of deforming a predefined template in shape modeling is also adopted in many other works.
For example, AtlasNet~\cite{AtlasNet} achieves better reconstruction quality by deforming multiple 2D planes. Deprelle \textit{et al.} \cite{ElemStruct} proposed to learn adaptive elementary structures that better fit the whole training data, and 3D-Coded~\cite{3d-coded} predicts correspondence for meshes via deforming the human template.

These methods demonstrate the ability for deep neural networks to deform one shape to another. 
However, existing works require either pre-defined or pre-generated templates (\textit{e.g.}, a 2D grid with fixed resolution, a learned set of elementary structures, or a 3D triangle mesh with fixed connectivity) to reconstruct the input point cloud or mesh.
In this paper, we show that the opposite direction, i.e., unfolding/deforming  arbitrary 3D point clouds to 2D regular grids of arbitrary resolution, is also applicable,  which has never been explored before.

\begin{figure*}[t]
	\centering
	\includegraphics[width=1\linewidth]{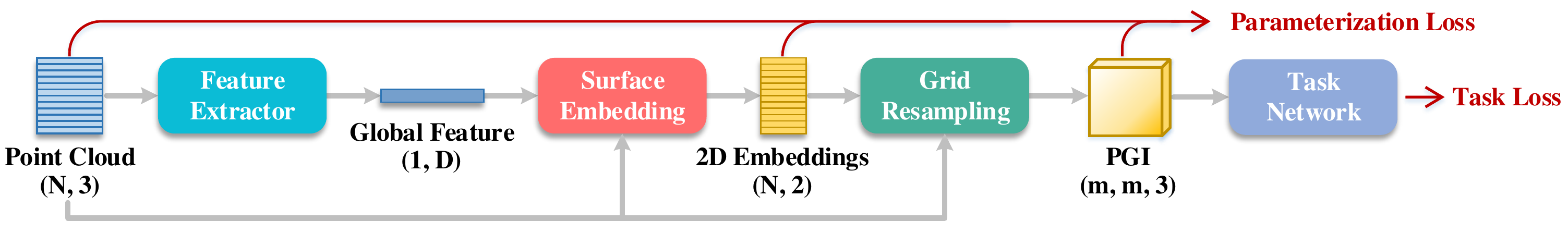}
	\caption{ParaNet consists of feature extractor and two core modules, namely surface embedding and grid resampling.
		The feature extractor computes a global feature for the input 3D point cloud. 
		Using the global feature, the surface unfolding module maps each 3D point to a unique 2D point in a square.
		Then the grid resampling module generates a geometry image, which can be fed into a CNN-based task network.
		The PGI generation is unsupervised and only regularized by geometric constraints denoted as the parameterization loss. 
		It can also be jointly optimized with the specific task loss.}
	\label{Fig03}
\end{figure*}

\begin{figure}[t]
	\begin{center}
		\includegraphics[width=1\linewidth]{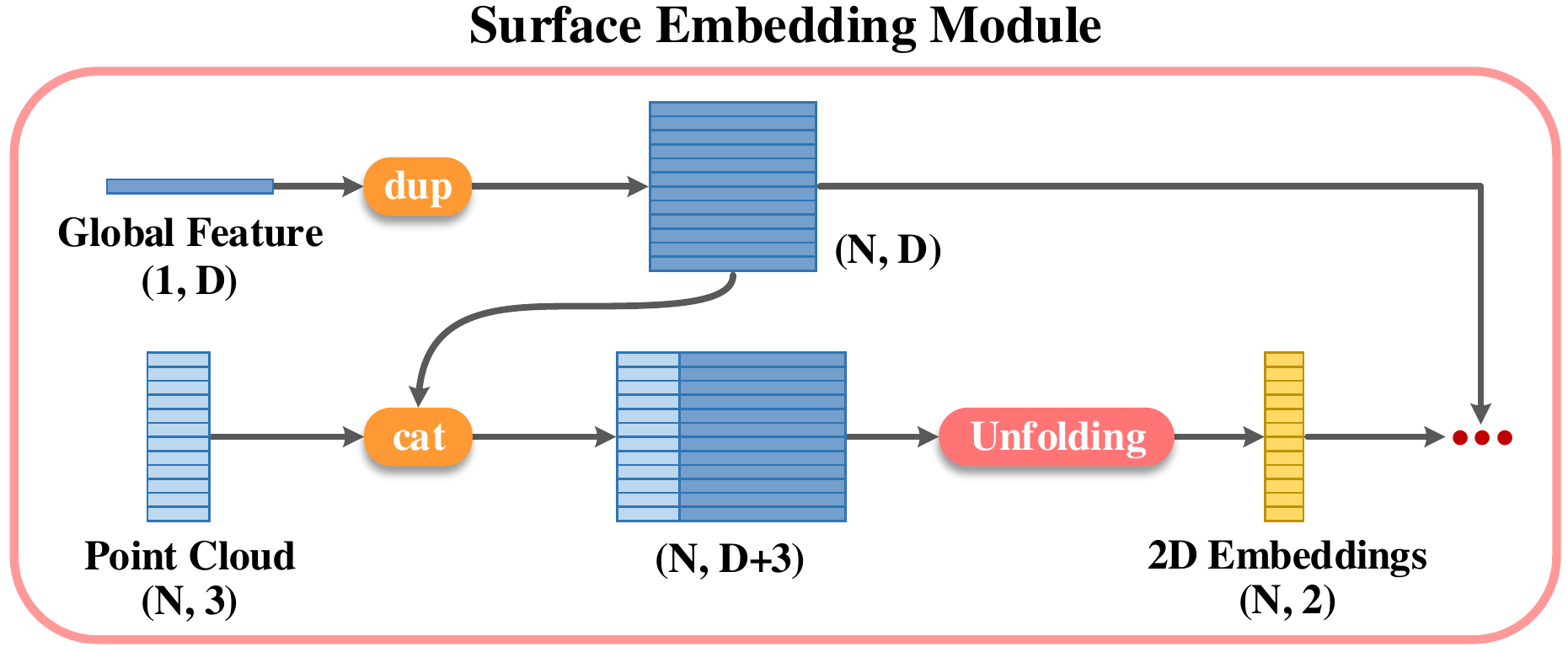}
	\end{center}
	\caption{The surface embedding module assigns each 3D point $\mathbf{p}_i$ a unique pair of 2D coordinates $(u_i,v_i)$, where $0\leq u_i,v_i\leq 1$.
	}
	\label{fig:embedding}
\end{figure}

\section{Proposed Method}
\label{metholology}

\subsection{Overview} \label{overview}
This work explores the potential to encode irregular 3D geometry in regular 2D grids and then apply conventional CNN architectures to perform specific point cloud processing tasks. Given a point cloud $\mathbf{P} \in \mathbb{R}^{N \times 3}$ with $N$ 3D points $\mathbf{p}_i=(x_i, y_i, z_i)$, where $i=1, 2, ..., N$, we construct a completely regular representation structure $\mathbf{G} \in \mathbb{R}^{m \times m \times 3}$, dubbed as a point geometry image (PGI), capturing the Cartesian coordinates of spatial points in the color channels of a 2D image. The generated PGI can be directly reshaped into a point cloud $\mathbf{Q} \in \mathbb{R}^{M \times 3}$, where $M = m \times m$. In practice, we always require $M>N$ so that the original points are redundantly represented in the PGI, reducing the loss of geometric information effectively.

The overall workflow of ParaNet is illustrated in Fig.~\ref{Fig03}. We start by extracting the global feature $\mathbf{c} \in \mathbb{R}^D$ from the input point cloud $\mathbf{P}$, and then generate a PGI in two stages. In the first stage, the surface embedding module maps each 3D spatial point to a 2D point over a unit square domain. In the second stage, the grid resampling module produces a regular point geometry image by either grid resampling or a differentiable relaxation of the nearest grid neighbor resampling. In our implementation, we employ the popular DGCNN \cite{DGCNN} as the backbone point feature extractor.

Thanks to the regular structure, PGIs can be fed into the subsequent CNN-based downstream task networks. In the paper, we demonstrate the potential of PGIs on classification (Sec.~\ref{sec:classification}) and upsampling (Sec.~\ref{sec:upsampling}).  \textit{Please refer to the Supplementary Material for the detailed discussions on the differences between our PGI and the classic GI representation.}

\subsection{Surface Embedding Module} \label{SUM}
The surface embedding module applies the unfolding operator twice to map each 3D point to a unique location in a unit square. See Fig.~\ref{fig:embedding}.
Specifically, we first duplicate the global feature $N$ times and arrange them in a row-wise matrix $\mathbf{C} \in \mathbb{R}^{N \times D}$.
Concatenating $\bf C$ with the input Cartesian coordinates $\mathbf{P}$, we obtain the initial embeddings $\mathbf{H}_1 \in \mathbb{R}^{N \times (D+3)}$. 
We then apply a four-layer perceptron to the rows of $\mathbf{H}_1$ 
to generate intermediate 2D embeddings $\mathbf{E}_1 \in \mathbb{R}^{N \times 2}$.

\begin{figure}
	\centering
	\includegraphics[width=1\linewidth]{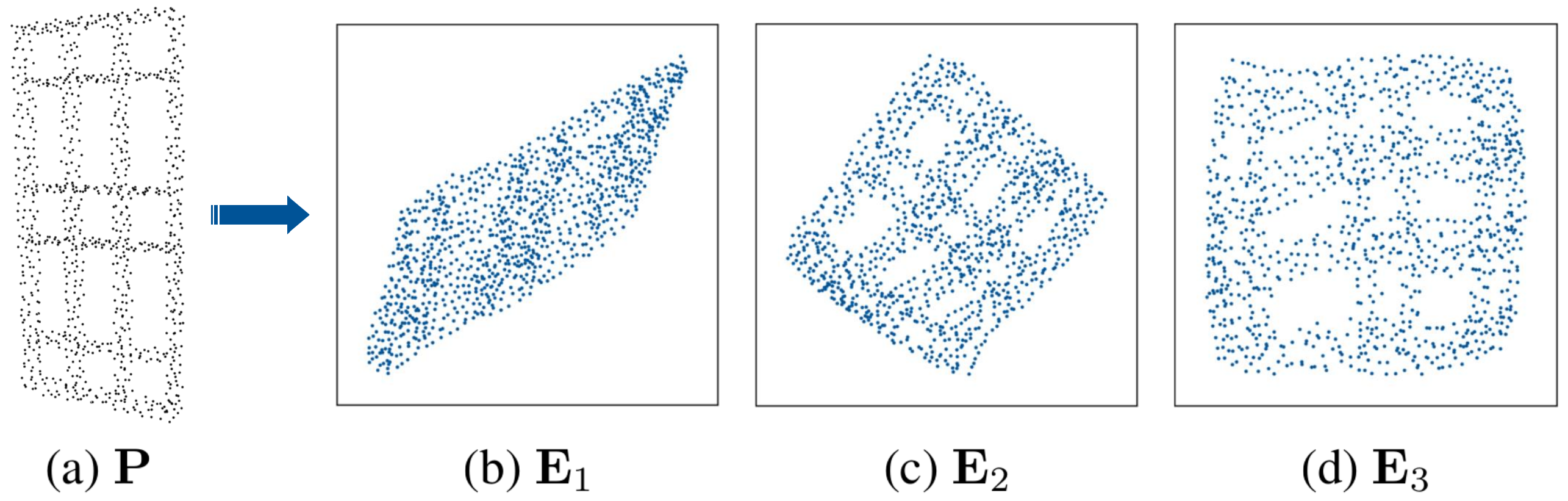}
	\caption{The unfold operators can be applied multiple times like function composition. (a) shows the input point cloud $\bf P$, and (b), (c) and (d) show the 2D embedded results of unfolding once, twice and three times, respectively. 
	}
	\label{fig:unfoldtwice}
\end{figure}

We apply the second unfolding operator in a similar fashion: we concatenate $\mathbf{E}_1$ with $\mathbf{C}$ to form $\mathbf{H}_2 \in \mathbb{R}^{N \times (D+2)}$,
and then feed $\mathbf{H}_2$ into another four-layer perceptron yields the final 2D embeddings $\mathbf{E} \in \mathbb{R}^{N \times 2}$. 

Putting it all together, we can express the surface embedding procedure as 
\begin{equation} \label{Euation-01}
	\mathbf{E} = \sigma(f_2([f_1([\mathbf{P};\mathbf{C}]);\mathbf{C}])),
\end{equation}
where $[\cdot;\cdot]$ denotes channel-wise concatenation between two feature matrices, $f_1(\cdot)$ and $f_2(\cdot)$ are two separate unfolding units, and $\sigma$ is the sigmoid activation function to normalize the embedded 2D points in the range of $[0, 1]$. The 2D embeddings after various unfold operators are visualized in Fig.~\ref{fig:unfoldtwice}. We observe that two-time unfolding leads to satisfactory results.

For convenience, we denote the $i$-th point contained in $\mathbf{E}$ as $\mathbf{e}_i=(u_i, v_i)$. It is worth mentioning that there exist one-to-one correspondence between the input point cloud $\mathbf{P}$ and the 2D embeddings $\mathbf{E}$, i.e., each 3D point  $\mathbf{p}_i \in \mathbb{R}^3$ is mapped to a unique 2D position $\mathbf{e}_i \in \mathbb{R}^2$.

%In the initial status, 
To prevent the embedded 2D points $\mathbf{e}_i$ from being highly clustered,  we design a simple repulsion loss to stretch the point distribution by imposing punishment on clustered points until they are separated by a distance threshold $T$.
Mathematically, we formulate the repulsion loss $\ell_{rep}(\cdot)$ for every point $\mathbf{e}_i$ as 
\begin{equation} \label{Euation-02}
	\begin{split}
		\ell_{rep}(\mathbf{e}_i)=&
		\begin{cases}
			0, & \text{if~$d_i \ge T$}\\
			-\log(d_i + 1 - T), & \text{otherwise}, \\
		\end{cases}\\
		&\text{where } d_i=\min_{\mathbf{e}_j\in \mathbf{E}\backslash\{\mathbf{e}_i\}}	\|\mathbf{e}_i-\mathbf{e}_j\|_2.
	\end{split}
\end{equation}
In our implementation we empirically set the distance threshold $T=\frac{1}{m-1}$. 

We define the objective function for the surface embedding module as: 
\begin{equation} \label{Eq-03}
	\mathcal{L}_{rep} = \frac{1}{N} \sum_{i=1}^N \ell_{rep}(\mathbf{e}_i).
\end{equation}
We observe that by minimizing $\mathcal{L}_{rep}$, the surface embedding module is capable of unfolding 3D surfaces onto a unit square. However, the embedded 2D points are not likely on the grid positions of the regular image lattice. Therefore, we need a resampling module (Sec.~\ref{GRM}) to produce a regular geometry image.

\subsection{Grid Resampling Module} \label{GRM}

\begin{figure}[t] 
	\begin{center}
		\includegraphics[width=1\linewidth]{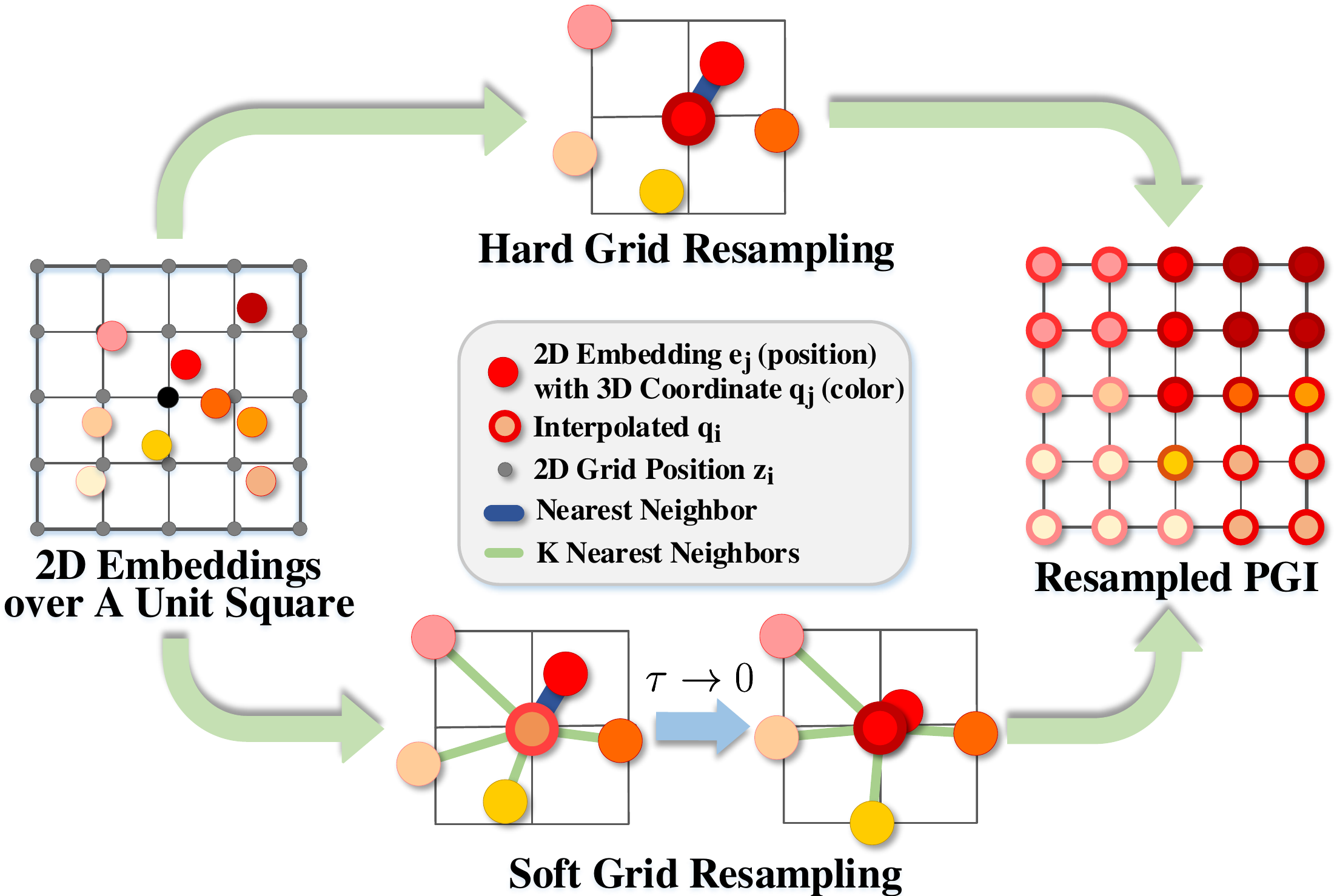}
	\end{center}
	\caption{The grid resampling module converts 2D parameters $\{(u_i,v_i)\}_{i=1}^N$ to an $m\times m$ geometry image. It supports both hard and soft resampling. The former maps each 2D sample to the nearest grid point and the latter adopts distance-weighted interpolation to assign the values for grid points.}
	\label{fig:grm}
\end{figure}

The grid resampling module consumes the irregularly-arranged 2D embeddings $\mathbf{E} \in \mathbb{R}^{N \times 2}$ and produces a regular PGI $\mathbf{G} \in \mathbb{R}^{m \times m \times 3}$ where $m$ is the user-specified resolution. 

There are two common strategies in resampling, as shown in Fig~\ref{fig:grm}. One is to drag each 2D sample $\mathbf{e}_i$ to the nearest pixel, and the other is to compute the pixel value using bilinear interpolation. Here, we denote the two strategies as hard grid resampling and soft grid resampling, respectively.  

\textbf{Hard Grid Resampling}. We start by constructing a canonical 2D grid uniformly defined on a unit square domain, denoted as $\mathbf{Z} \in \mathbb{R}^{M \times 2}$ containing $M$ points. Each row of matrix $\mathbf{Z}$ corresponds to a certain 2D point $\mathbf{z}_i$ within an $m \times m$ gridding structure. 
Suppose that, in the point set $\mathbf{E}$, the nearest neighbor of $\mathbf{z}_i$ is $\mathbf{e}_j$.
Based on the exact row-wise matching relations between $\mathbf{P}$ and $\mathbf{E}$, we can deduce a 2D array $\mathbf{Q} \in \mathbb{R}^{M \times 3}$ whose $i$-th entry $\mathbf{q}_i \in \mathbb{R}^3$ is the $j$-th entry of $\mathbf{P}$. We formulate the hard resampling as follows:
\begin{equation} \label{Eq-04}
	\mathbf{q}_i = \mathbf{p}_{j^*} \text{, where }j^* = \mathop{\arg\min}_{j \in \{1,2,\cdots,N\}} \|\mathbf{z}_i- \mathbf{e}_j\|_2.
\end{equation}
After that, we reshape the $M \times 3$ array $\mathbf{Q}$ into an $m \times m \times 3$ PGI $\mathbf{G}$. As two mapped samples $\mathbf{e}_i$ and $\mathbf{e}_j$ could be cluttered, there is a possibility that some 3D points are not included in $\mathbf{G}$. This problem can be reduced by increasing the PGI resolution $m$. With a sufficiently high resolution, PGI is guaranteed to include all points of the input point cloud, resulting a lossless representation.

\textbf{Soft Grid Resampling}. Notice that the hard grid resampling workflow is non-differentiable, which hinders end-to-end optimization. To overcome this limitation, we further provide an annealing strategy to produce PGIs in a differentiable manner. In soft resampling, we model $\mathbf{q}_i$  as a weighted average of the input points $\mathbf{P}$. For grid point $\mathbf{z}_i$,
we select its $K$ nearest neighbors from $\mathbf{E}$, which are denoted as $\{\mathbf{e}_k\}_{k=1}^K$, and their corresponding 3D points  $\{\mathbf{p}_k\}_{k=1}^K$. 
% $\mathcal{N}_\mathbf{E}(\mathbf{z}_i)=\{ s_{ik}\}_{k=1}^K$, 
We then determine a set of weights using the distances between $\mathbf{z}_i$ and its $K$ neighbors:
\begin{equation} \label{Eq-06}
	\varpi_{ik} = \frac {\exp(-d_{ik}/|\tau|)} {\sum_{j=1}^{K}\exp(-d_{ij}/|\tau|)},
\end{equation}
where $d_{ik}=\|\mathbf{z}_i- \mathbf{e}_k\|_2$, and $\tau$ is a temperature coefficient. 
We express the $i$-th entry of $\mathbf{Q}$ as 
\begin{equation} \label{Eq-07}
	\mathbf{q}_i = \sum_{k=1}^K \varpi_{ik}  \mathbf{p}_k.
\end{equation}

During annealing, the distribution of weights gradually converges to the Kronecker delta function with $\tau \rightarrow 0$, approximating the nearest neighbor point. We can easily achieve this by adding a parameter regularization term:
\begin{equation} \label{Eq-08}
	\mathcal{L}_{ann} = |\tau|
\end{equation}

Combining Eq. (\ref{Eq-03}) and Eq. (\ref{Eq-08}), we obtain the objective function for the unsupervised ParaNet:
\begin{equation}
	\mathcal{L}_{para} = \alpha  \mathcal{L}_{rep} + \beta  \mathcal{L}_{ann}.
	\label{eqn:joint}
\end{equation}

Note that annealing is dropped in hard grid resampling.

\begin{figure}[t] 
	\begin{center}
		\includegraphics[width=1\linewidth]{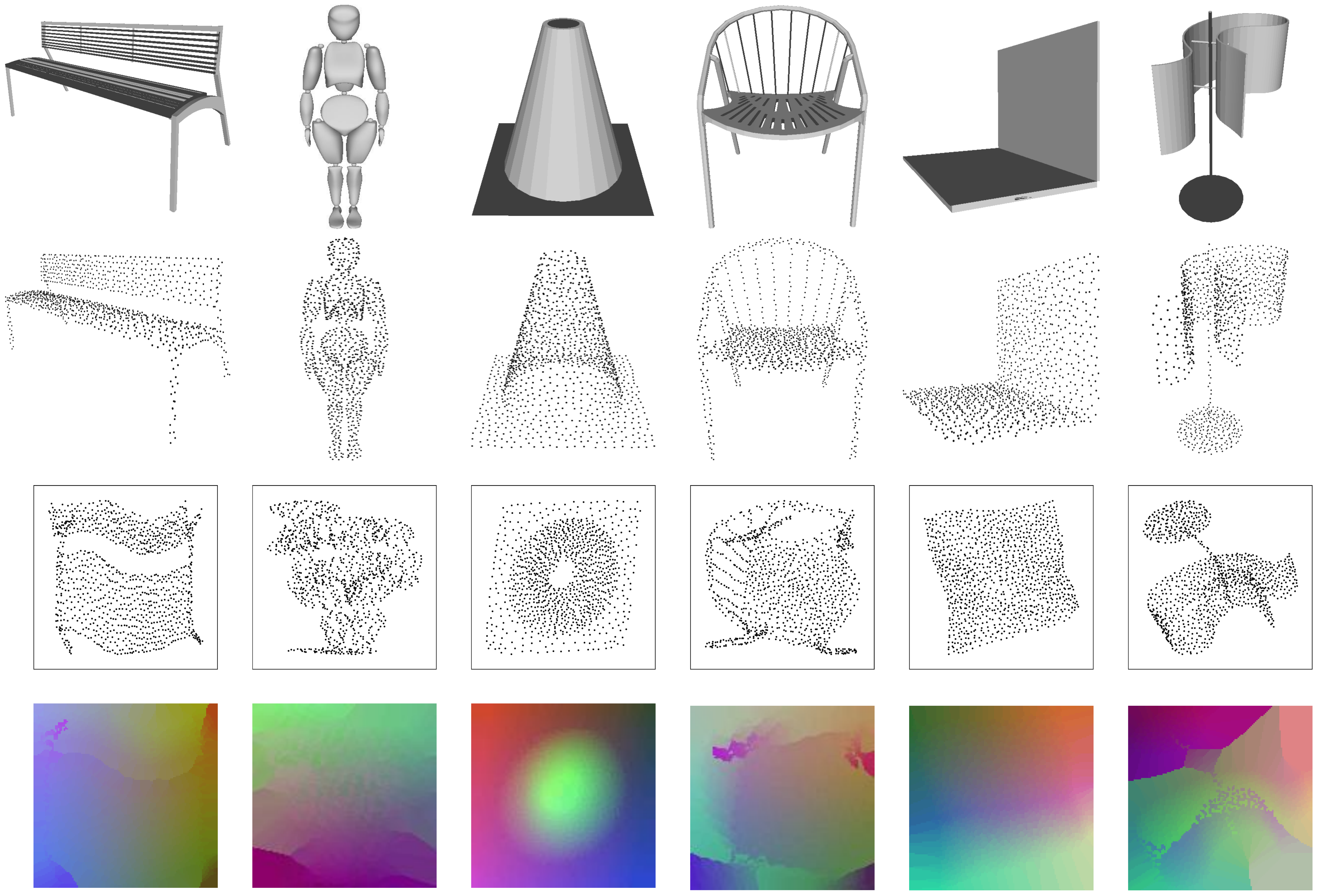}
	\end{center}
	\caption{PGI results.
		Row 1: 3D meshes.
		Row 2: Point clouds $\mathbf P$.
		Row 3: 2D embedded results $\mathbf{E}_2$.
		Row 4: Point geometry images of resolution $128 \times 128$.
		From left to right, the Chamfer distance errors ($10^{-8}$) are 1.71, 1.28, 1.89, 1.04, 1.72 and 1.78 respectively.
		See \textit{Supplementary Material} for more results.
	}
	\label{Figure-example}
\end{figure}

Fig.~\ref{Figure-example} shows some visual examples by running ParaNet over input shapes after sufficient iterations. We can observe that the PGIs accurately represent the original point clouds while maintaining satisfactory smoothness.

\section{Experimental Results}
\label{sec:results}

\subsection{Training Strategies} \label{sec_4_1}
As an unsupervised learning approach, ParaNet enables flexible and customized optimization schemes. In order to fit the specific characteristics and requirements of different application scenarios, we investigated three types of training strategies: 1) joint optimization; 2) offline independent optimization; and 3) online independent optimization. Note that we employ soft grid resampling for joint optimization, and hard grid resampling for independent optimization. 

\textbf{Joint Optimization (JO).} ParaNet and the downstream task network are jointly trained with a combined objective function involving the parameterization loss $\mathcal{L}_{para}$ and the task-specific loss $\mathcal{L}_{task}$. We pre-train the ParaNet before connecting it with the task network to stabilize and speed up the joint training process. The JO strategy benefits from the task-oriented learning paradigm. Thanks to the implicit guidance of the task network, the generated PGIs can be automatically adjusted to suit downstream applications. 

\textbf{Offline Independent Optimization (Off-IO).} We start by pre-training  ParaNet on the whole training set, and then freeze its parameters before connecting it with the downstream task network. Accordingly, we only exploit the task-specific loss $\mathcal{L}_{task}$ to supervise the overall framework. We observe that the independent optimization strategy usually leads to PGIs with higher representation accuracy, while sacrificing the possibility of adaptively adjusting the geometry parameterization process.

\textbf{Online Independent Optimization (On-IO).} We separately optimize ParaNet for every input shape with adequate iterations to obtain a nearly lossless representation during both training and testing. We can achieve this by generating an accurate PGI for every single input shape over the training and testing sets. The generated PGIs are cached and further fed into the downstream task network. Obviously, the On-IO strategy is much slower than the offline strategy, but it produces highly-precise representation of input point clouds with little information loss. In practice, On-IO can be greatly accelerated by feeding batch-wise inputs.

\subsection{Shape Classification} \label{sec:classification}

\begin{figure}[t] 
	\begin{center}
		\includegraphics[width=1\linewidth]{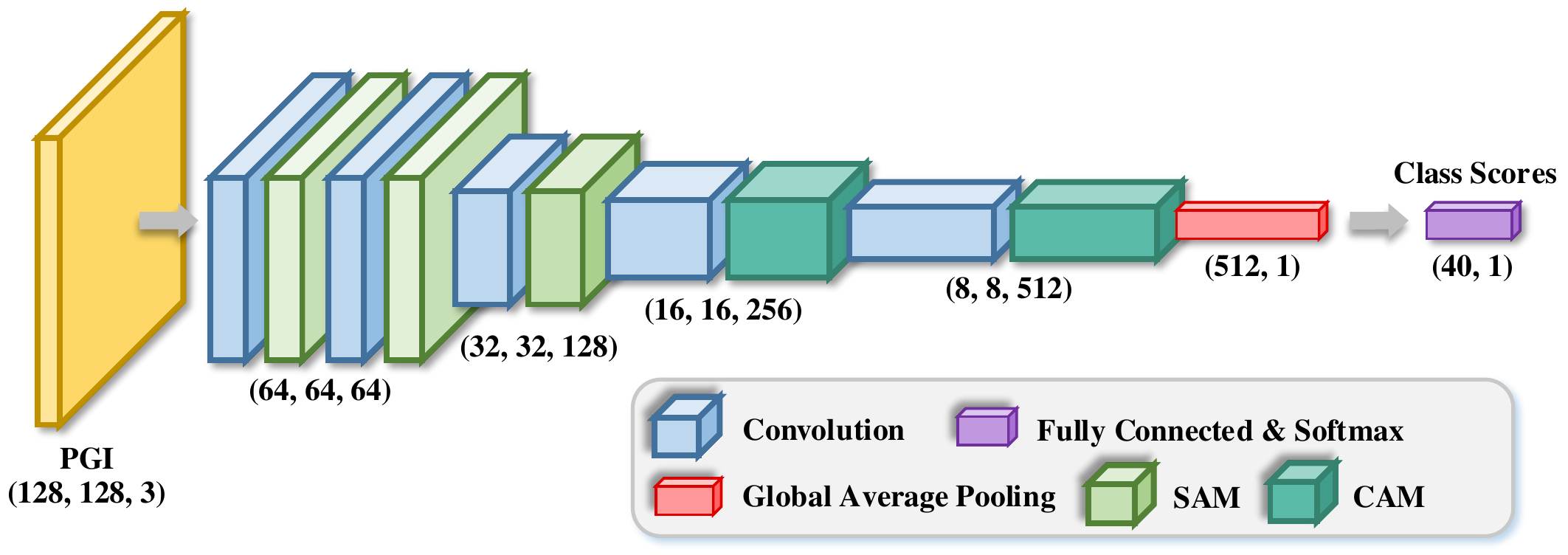}
	\end{center}
	\caption{ParaNet fed into a classification network.}
	\label{fig:classification_net}
\end{figure}

\textbf{Task network}. Image classification has been adequately addressed in numerous CNN-based architectures. In our shape classification task, we adopted the commonly-used ResNet-18 \cite{RegCNN-7} architecture, which consists of five convolution layers, as illustrated in Fig.~\ref{fig:classification_net}. Considering that the original point cloud is redundantly represented in the PGI, we applied three spatial attention modules (SAM) to highlight informative regions, and another two channel attention modules (CAM) were equipped to boost feature learning. Afterwards, the extracted feature maps were squeezed into a one-dimensional feature vector by global average pooling, and passed through a fully-connected layer followed by Softmax to predict the class probabilities. We employed cross-entropy as the task loss to supervise the task network. Notice that we did not utilize the ImageNet \cite{RegCNN-1} pretrained parameters for fair comparisons.

\begin{table}[!t]
	\renewcommand\arraystretch{1}
	\caption{Classification accuracy on ModelNet40.}
	\begin{center}
		\setlength{\tabcolsep}{1.0mm}{
			\begin{tabular}{ c | c | c }
				\hline\hline
				Method & Input & Accuracy (\%) \\
				\hline
				VoxNet \cite{VoxNet} & voxels & 85.9  \\
				KD-Net \cite{KD-Net} & 1024 points & 91.8  \\
				PointNet \cite{PointNet} & 1024 points & 89.2  \\
				PointNet++ \cite{PointNet++} & 5000 points + normal & 91.9  \\
				3D-GCN \cite{3D-GCN} & 1024 points & 92.1 \\
				SpiderCNN \cite{SpiderCNN} & 1024 points + normal & 92.4  \\
				PointConv \cite{PointConv} & 1024 points + normal & 92.5  \\
				A-CNN \cite{A-CNN} & 1024 points & 92.6  \\
				KPConv \cite{KPConv} & 6800 points & 92.9  \\
				DGCNN \cite{DGCNN} & 1024 points & 92.9  \\
				\hline
				ParaNet-SC (JO) &  1024 points &  \textbf{93.1} \\
				ParaNet-SC (Off-IO) &  1024 points & 92.7  \\
				ParaNet-SC (On-IO) &  1024 points & 92.9  \\
				\hline\hline
		\end{tabular}}
	\end{center}
	\label{table-01}
\end{table}

\textbf{Development protocols}. We evaluated the shape classification performance of our framework over the ModelNet40 \cite{ModelNet} benchmark dataset, which contains $12311$ CAD models from $40$ man-made object categories. Following the official train-test partition, we employed $9843$ shapes for training and the rest $2468$ for testing. We used Poisson Disk Sampling to generate $1024$ points from the original mesh. Common data augmentation techniques including random point dropout and jittering point positions by Gaussian noises were employed to enrich the training data and boost the generalization ability of the learning model. For parameters in Eq.~\ref{eqn:joint}, we empirically set the parameters $\alpha=1$ and $\beta=0.1$.

\textbf{Comparison with state-of-the-art methods}. We report the performances of the proposed framework under different training strategies in Table \ref{table-01}, which includes $10$ classic and latest state-of-the-art competitors for comparisons. As shown in Table \ref{table-01}, our ParaNet  model under the JO strategy achieves the highest classification accuracy compared with other rasterization- and point-based frameworks. An interesting observation is that, although the Off-IO and On-IO schemes typically produce more accurate PGI representations, the JO scheme shows superiority benefiting from the end-to-end task-oriented learning paradigm. 

\textbf{Ablation studies}. We conducted ablation studies to explore the effectiveness and potential of the proposed PGI representation. The results of different model variants of the ParaNet-JO are reported in Table \ref{table-02}. To demonstrate the necessity of representing raw point clouds as PGIs, we removed the ParaNet from the full pipeline and directly and randomly reshape the input point cloud from a 2D array to the form of a color image, denoted as ParaNet-SC-random. Under this circumstance, the performance significantly drops from $93.1\%$ to $45.6\%$, which reveals that it is non-trivial to apply standard convolutions to process irregular and unstructured point clouds. Then, we designed a naive baseline CNN architecture serving as the downstream task network to form the method ParaNet-SC-baseline. Details of the baseline design can be found in the \textit{Supplemental Material}. This simple variant also achieves favorable performance of $91.9\%$. When we replaced the backbone network in the baseline with a more powerful ResNet18 \cite{RegCNN-7} encoder, the classification accuracy for ParaNet-SC-ResNet18 increases to $92.6\%$. Moreover, we explored whether the success of image attention mechanisms can be extended to our task setting. Empirically, we observe performance decrease when the SAM or the CAM modules are removed from the full pipeline, which implies that our method enables direct adaptation of existing visual techniques that have been proven to be effective in image domain applications. We can expect to achieve further performance gains as more advanced 2D techniques are introduced to cooperate with ParaNet. 

We further explored the impact of generating the PGIs with different resolutions. As discussed before, the original point cloud is redundantly represented in image pixels. Predictably, the information loss will decrease as the image resolution increases. Thus, we are able to flexibly adjust the representation accuracy at the cost of the representation redundancy. As reported in Table \ref{table-03}, low resolution of the parameterized PGI leads to degraded performance, since a noticeable number of points may be lost during grid resampling. When the image resolution is sufficiently large to accurately represent the point cloud, the performance gain vanishes accordingly.

\begin{table}[!t]
	\renewcommand\arraystretch{1}
	\caption{Quantitative comparisons among different variants under the JO strategy over the ModelNet40 dataset.}
	\begin{center}
		\setlength{\tabcolsep}{5mm}{
			\begin{tabular}{ c | c }
				\hline\hline
				Method & Accuracy (\%) \\
				\hline
				ParaNet-SC-random & 45.6  \\
				ParaNet-SC-baseline & 91.9 \\
				ParaNet-SC-ResNet18 & 92.6 \\
				ParaNet-SC \textit{w/o} CAM & 92.8 \\
				ParaNet-SC \textit{w/o} SAM & 93.0 \\
				ParaNet-SC (full) & \textbf{93.1} \\
				\hline\hline
		\end{tabular}}
	\end{center}
	\label{table-02}
\end{table}

\begin{table}[!t]
	\renewcommand\arraystretch{1.1}
	\caption{Impacts of different resolution of PGIs.}
	\begin{center}
		\setlength{\tabcolsep}{5mm}{
			\begin{tabular}{ c | c | c }
				\hline\hline
				Resolution & Accuracy (\%) & CD Errors\\
				\hline
				$32  \times  32$ & 91.6 & $1.14 \times 10^{-3}$ \\
				$48  \times  48$ & 92.5 & $2.12 \times 10^{-4}$ \\
				$64  \times  64$ & 93.0 & $4.49 \times 10^{-5}$ \\
				$96  \times  96$ & \textbf{93.1} & $7.38 \times 10^{-6}$ \\
				$128 \times 128$ & \textbf{93.1} & $\mathbf{2.32 \times 10^{-6}}$ \\
				\hline\hline
		\end{tabular}}
	\end{center}
	\label{table-03}
\end{table}

\subsection{Point Cloud Upsampling} \label{sec:upsampling}

\begin{figure}[t] 
	\begin{center}
		\includegraphics[width=1\linewidth]{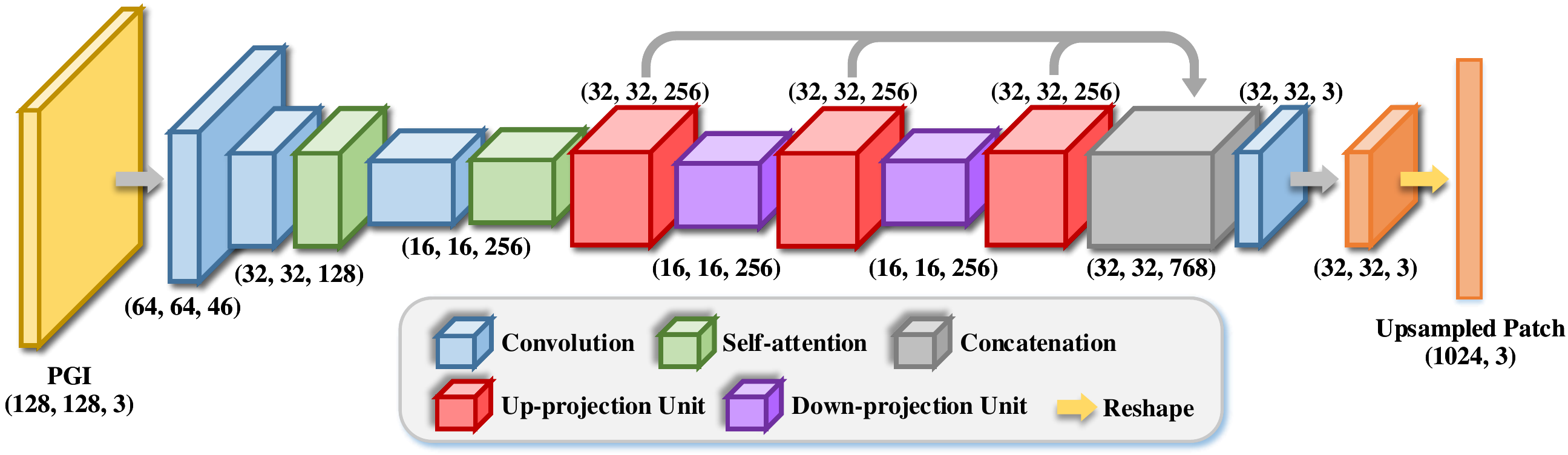}
	\end{center}
	\caption{ParaNet fed into a upsampling network.}
	\label{fig:upsampling_net}
\end{figure}

\begin{figure*}[t]
	\begin{center}
		\includegraphics[width=1\linewidth]{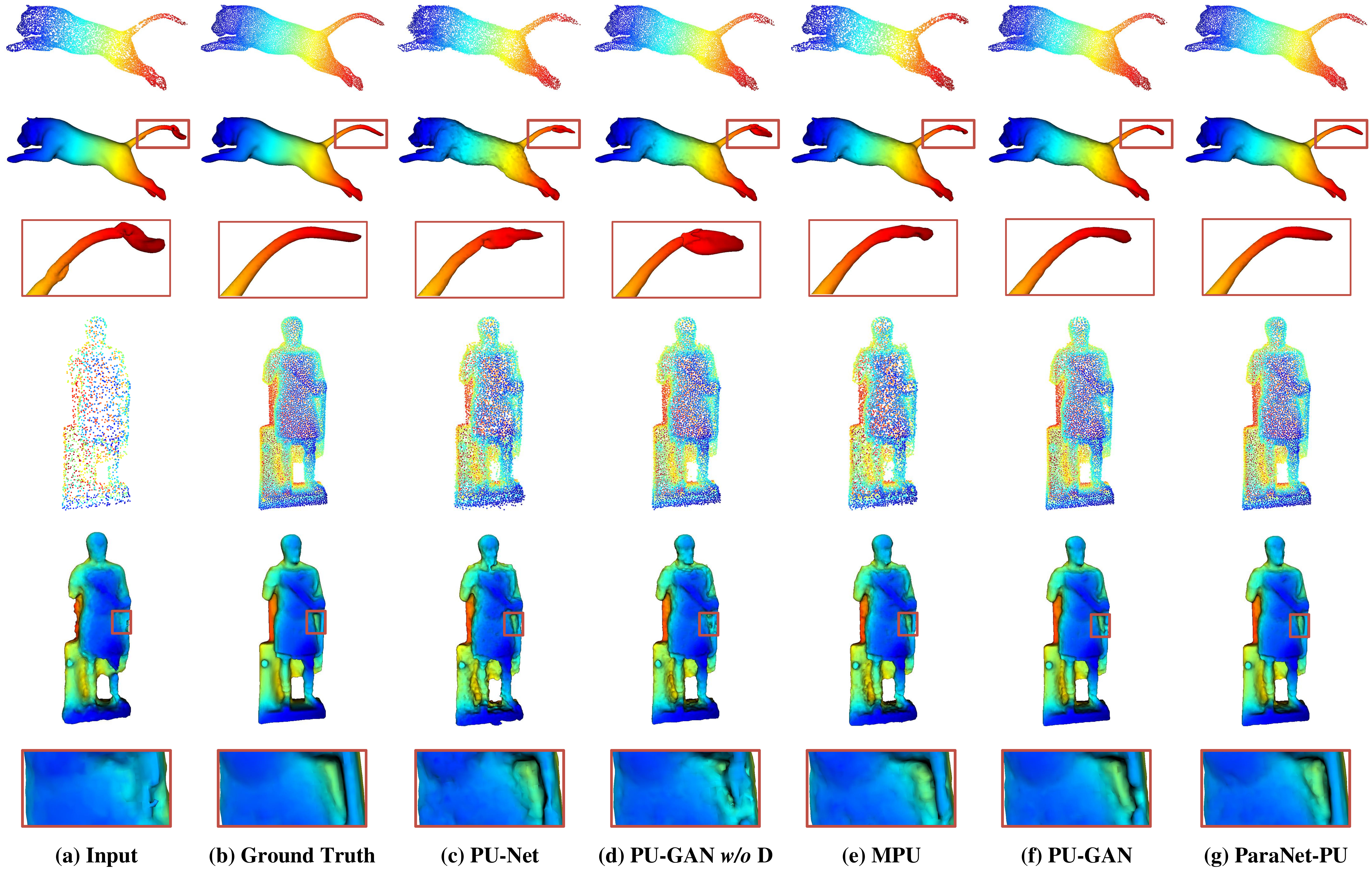}
	\end{center}
	\caption{Point cloud upsampling with a factor $4 \times$. To ease checking the result quality, we show the triangle meshes reconstructed from the upsampled point clouds. See \textit{Supplementary Material} for more results.}
	\label{visual-pu}
\end{figure*}

\textbf{Task network.} Extensive studies have been conducted for image super-resolution, which provide valuable insights for point cloud upsampling. We followed the common patch-based processing pipeline \cite{PU-GAN} to achieve $4 \times$ upsampling for non-uniform inputs. As illustrated in Fig.~\ref{fig:upsampling_net}, we first converted input patches into PGIs, which passed through the subsequent convolutional layers for feature embedding. The back projection mechanism \cite{BackProjection} was introduced to fully exploit the mutual dependencies of low- and high-resolution contents, and self-attention operations \cite{wang2018non} were introduced to encode global prior of the input patch. We utilized the Earth Mover's Distance (EMD) to supervise point cloud generation, and employed a uniformity constraint to enhance point distribution. Different from shape classification, the upsampling task requires highly-accurate representations with little information loss. Therefore, we employed the On-IO scheme for training. 

\textbf{Development protocols.} Following previous study \cite{PU-GAN}, we collected $147$ 3D mesh models from the released datasets of PU-Net \cite{PU-Net} and MPU \cite {MPU} as well as Visionair repository \cite{Visionair}, among which $120$ for training and $27$ for testing. We used Poisson disk sampling to sample $8192$ points as ground-truths, and randomly selected $2048$ points as inputs. The size of extracted patches is $256$, and the resolution of converted PGIs is $128 \times 128$. We employed three evaluation metrics: 1) point-to-surface (P2F) distance, 2) Chamfer distance (CD), and 3) Hausdorff distance (HD).

\textbf{Comparison with state-of-the-art methods}. We compared our pipeline, \textit{i.e.}, ParaNet-PU, with three deep point upsampling models by retraining their models on our dataset. Note that PU-GAN \cite{PU-GAN} additionally adopts adversarial learning, hence, we also report the performance of its point generator, \textit{i.e.}, PU-GAN \textit{w/o} D. As reported in Table \ref{table-ups}, our PGI-driven framework outperforms these point-based architectures. We provided some visual comparisons of upsampled point clouds and the reconstructed surfaces using \cite{kazhdan2013screened} in Fig. \ref{visual-pu}, from which we can observe that our results better restore detailed shape geometry and contain less noises that deviating from the underlying surface.

\begin{table}[!t]
	\renewcommand\arraystretch{1.1}
	\caption{Quantitative comparison of different methods for $4\times$ point cloud upsampling. The best results are highlighted in bold.}
	\begin{center}
		\setlength{\tabcolsep}{3.0mm}{
			\begin{tabular}{ c | c | c | c}
				\hline\hline
				Method & P2F & CD & HD \\
				& $(10^{-3})$ & $(10^{-3})$ & $(10^{-3})$ \\
				\hline
				PU-Net \cite{PU-Net} & 6.97 & 0.72 & 8.93  \\
				PU-GAN \textit{w/o D} \cite{PU-GAN} & 4.75 & 0.59 & 7.44  \\
				MPU \cite{MPU} & 3.93 & 0.49 & 6.06  \\
				PU-GAN %\textit{w/ D}
				\cite{PU-GAN} & 2.40 & 0.29 & 4.75  \\
				ParaNet-PU & \textbf{2.26} & \textbf{0.28} & \textbf{4.44}  \\
				\hline\hline
		\end{tabular}}
	\end{center}
	\label{table-ups}
	\vspace{-0.5cm}
\end{table}

\section{Conclusion}
We proposed an end-to-end deep neural network, \textit{i.e.}, ParaNet, to convert irregular 3D point clouds into a completely regular 2D representation modality, which is an ideal input of standard 2D convolutional neural networks. ParaNet can be trained either in an unsupervised manner or jointly with the specific downstream task network. We demonstrated the effectiveness of ParaNet on shape classification and point cloud upsampling. Extensive experiments demonstrate the superiority of our point cloud processing pipelines. Our method naturally bridges the gap between 2D CNNs and 3D point cloud processing. We believe it will open a door to apply well-developed 2D deep learning tools to 3D applications. In future studies, we would like to explore richer applications based on point geometry image for point cloud analysis.

%%%%%%%%%%%%%%%%%%%
\newpage
\onecolumn
\noindent\textbf{{\large Supplementary Materials}}

\section*{1. Differences Between the Proposed PGIs and the Classic GIs}
Both the proposed point geometry images (PGIs) and the classic geometry images (GIs)~\cite{GI} are able to represent 3D shapes with a completely regular structure, \textit{i.e.}, a color image. When the image resolution is sufficiently high, both can encode 3D geometry in a nearly lossless manner. Besides the above two common characteristics, the proposed PGIs are fundamentally different from the classic GIs in terms of generation and application. 

GIs are generated through mesh-based parameterization algorithms, which are usually formulated as either a non-linear optimization problem that minimizes area and/or angle distortions~\cite{DBLP:journals/tog/BommesZK09,DBLP:journals/cgf/LiuZXGG08,DBLP:journals/tog/SpringbornSP08,DBLP:journals/tog/ShefferLMB05} or a non-linear PDE that iteratively deforms the input 3D shape to a target shape with constant Gaussian curvature~\cite{DBLP:journals/cgf/ZhaoSLZYLWGG20,DBLP:journals/tvcg/JinKLG08}. GIs are typically judged by the overall angle and area distortion, and applied to GPU-accelerated rendering and texture mapping (\textit{e.g.}, in movies and video games). The mesh-based algorithms are able to compute high-quality GIs for 3D models with simple geometry and topology. However, increasing the complexity of geometry/topology often poses significant challenges to these methods. Furthermore, they only work for the manifold meshes with a single component, whereas the non-manifold models with multiple connected components are common in real-world applications. Last but not least, extending the mesh-based parameterization algorithms to point clouds is non-trivial due to their high reliance on the connectivity information provided by meshes. 

Our method is data-driven and can work for arbitrary manifold and non-manifold models with single or multiple components. Since ParaNet makes no use of connectivity, it is ideal for processing point clouds. As an end-to-end learning framework, it is trained in an unsupervised manner. Moreover, it can be combined and jointly optimized with downstream task networks. Since PGIs can be seamlessly coupled with conventional 2D CNN architectures, they are particularly useful for visual analysis tasks, such as classification, segmentation, super-resolution and recognition, in which 2D CNNs have already demonstrated great potentials. 

Due to these fundamental differences, we do not na\"{i}vely follow conventional GI metrics to measure PGIs. In our paper, we evaluate PGIs in terms of: 1) the information loss between the input point cloud and the PGI-induced point cloud; and 2) the performance of specific downstream applications when applying 2D CNNs to the generated PGIs.

In Figure \ref{PGIs-2}, we provide more examples of the 2D embeddings and PGIs. The displayed models belong to the ModelNet40 \cite{ModelNet} repository. We observe that ParaNet works well for diverse object categories.

\begin{figure*}
	\centering
	\includegraphics[width=0.95\linewidth]{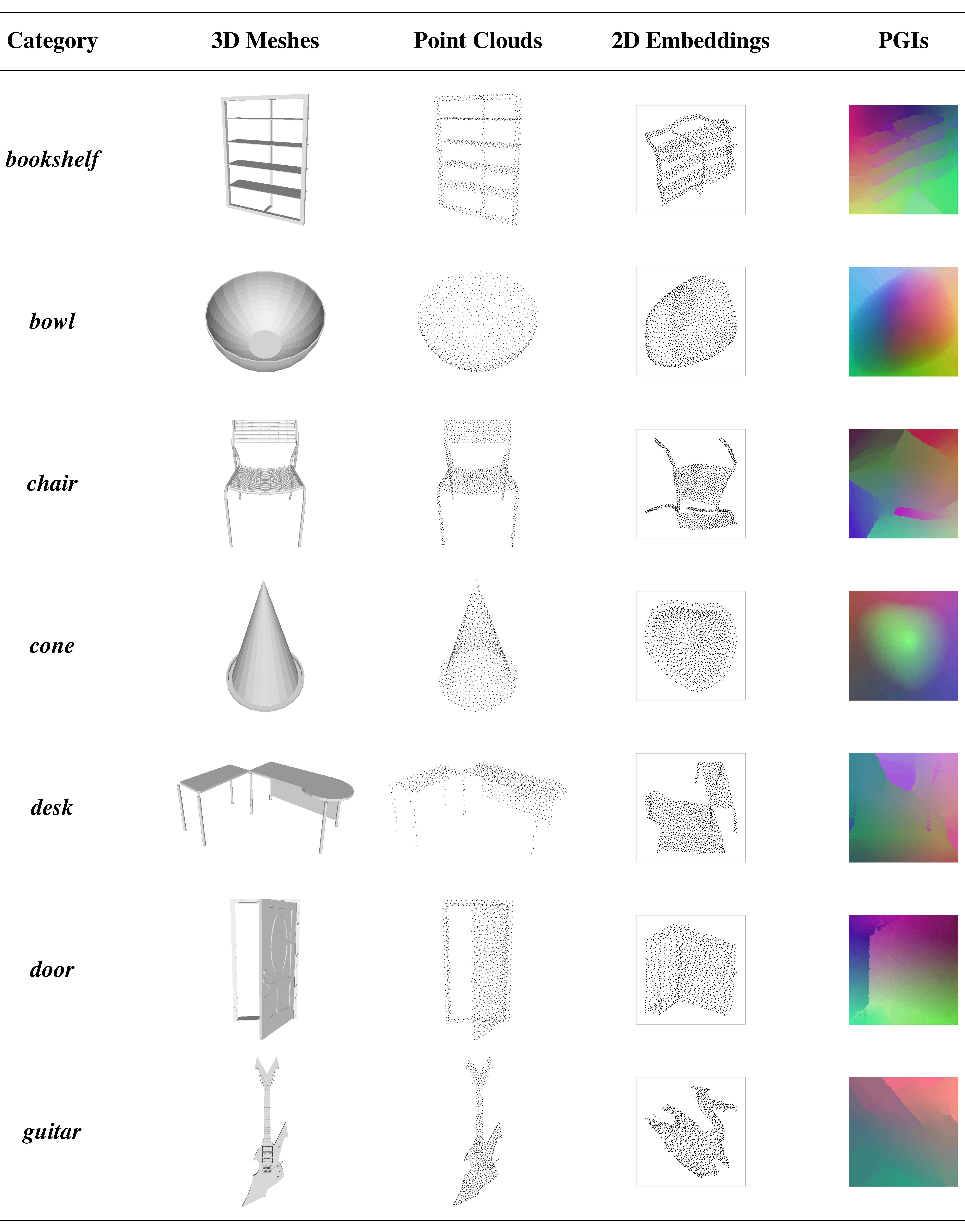}
	\label{PGIs-1}
\end{figure*}

\begin{figure*}
	\centering
	\includegraphics[width=0.95\linewidth]{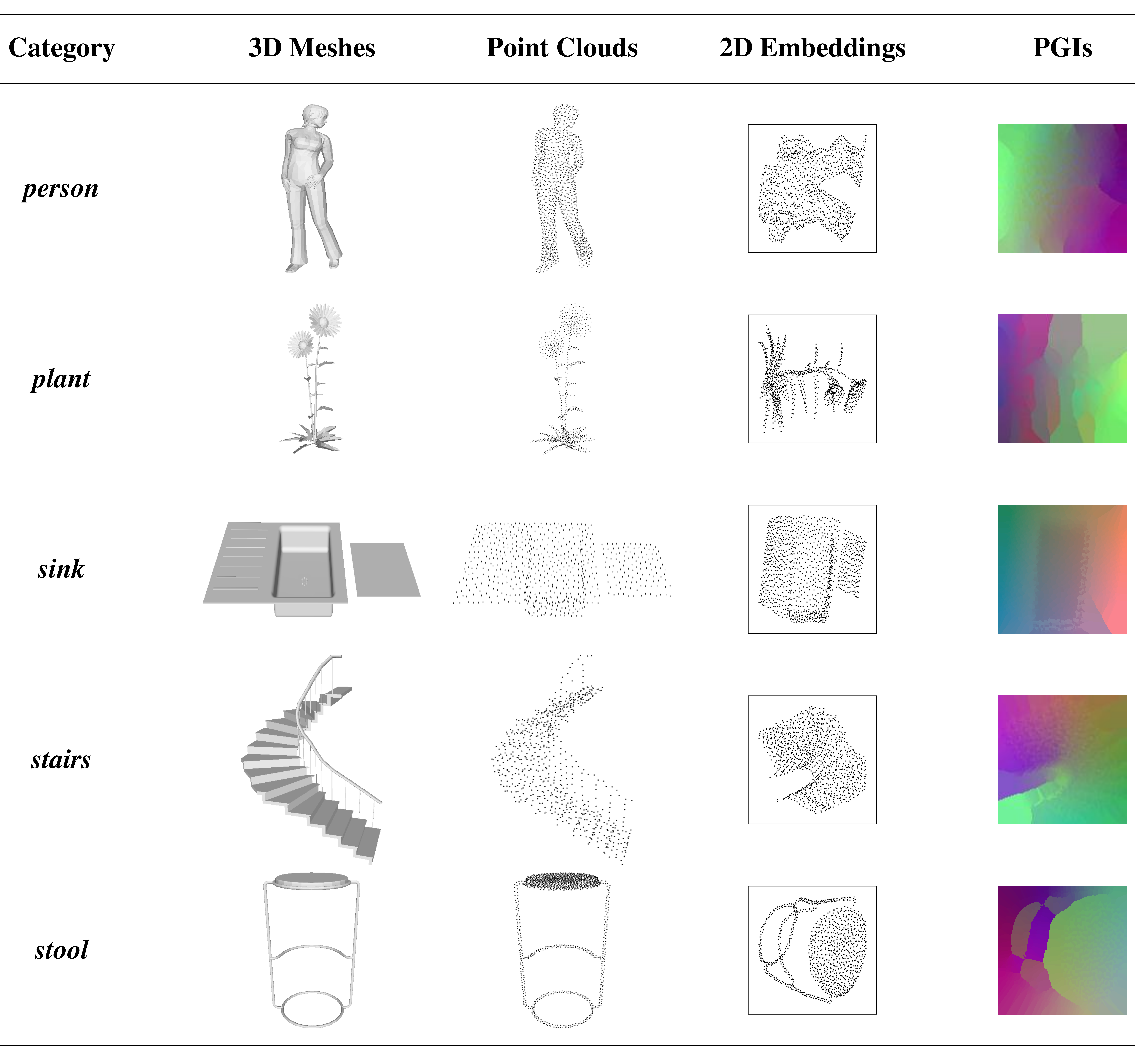}
	\caption{More results of the 2D embeddings and PGIs produced by ParaNet.}
	\label{PGIs-2}
\end{figure*}

\section*{2. Details of Network Architectures}

\subsection*{2.1 ParaNet}
In our implementation, we employed the popular DGCNN \cite{DGCNN} architecture for feature extraction of input point clouds. The extracted global feature for an input point cloud is a $512$-dimensional feature vector, \textit{i.e.}, $\mathbf{c} \in \mathbb{R}^{512}$. In the surface embedding module, each unfolding operator was implemented as a shared four-layer perceptron. The dimensions of the first and the second unfolding operator are $\{ 512+3, 256, 128, 64, 2 \}$ and $\{ 512+2, 256, 128, 64, 2 \}$, respectively. For the annealing regularization term in soft grid resampling, we initiated the learnable temperature coefficient as $\tau = 10^{-5}$. Besides, the number of nearest neighbors selected for each grid point was set as $K=5$. In practice, we discover that the soft grid resampling module is not sensitive to the value of $K$ due to the effective annealing process.

\subsection*{2.2 ParaNet-SC: Downstream Task Network for 3D Shape Classification}

\begin{table}[!t]
	\renewcommand\arraystretch{1.1}
	\caption{Components and outputs of ParaNet-SC.}
	\begin{center}
		\setlength{\tabcolsep}{3mm}{
			\begin{tabular}{ c | c }
				\hline\hline
				Components & Outputs ([Channels, Height, Width]) \\
				\hline
				PGI    & [3, 128, 128] \\
				Conv-1 & [64, 64, 64]  \\
				SAM-1  & [64, 64, 64]  \\
				Conv-2 & [64, 64, 64]  \\
				SAM-2  & [64, 64, 64]  \\
				Conv-3 & [128, 32, 32] \\
				SAM-3  & [128, 32, 32] \\
				Conv-4 & [256, 16, 16] \\
				CAM-4  & [256, 16, 16] \\
				Conv-5 & [512, 8, 8]   \\
				CAM-5  & [512, 8, 8]   \\
				Global Average Pooling    & [512, 1, 1]   \\
				Fully Connected Layer     & [40]          \\
				\hline\hline
		\end{tabular}}
	\end{center}
	\label{Sup-Table-1}
\end{table}

The ParaNet-SC framework tailored for 3D shape classification adopted the ResNet18 \cite{RegCNN-7} architecture, which has been widely applied to a variety of image domain visual applications, as the backbone feature extractor. Following common notations, the standard ResNet18 feature encoder can be partitioned and packed into five sequential convolution stages, denoted as: $\lbrace$Conv-1, Conv-2, Conv-3, Conv-4, Conv-5$\rbrace$. Note that we dropped the max-pooling layer in the Conv-1 stage to enlarge the feature resolution.

Considering that the generated PGIs usually represent the original point clouds in a highly-redundant manner, it is necessary to deal with the information redundancy issue. Thus, we introduced the popular attention mechanism, as proposed in \cite{CBAM}, which has already gained remarkable success in numerous image/video processing tasks. In \cite{CBAM}, a spatial attention module (SAM) is applied to aggregate inter-spatial feature dependency and suppress the less informative image parts, and a channel attention module (CAM) is applied to exploit inter-channel relationships and emphasize the most discriminative feature subspace. Intuitively, the combination of SAM and CAM enables the network to learn \textit{where} and \textit{what} to emphasize or suppress, effectively enhancing the representation capability of the learned convolutional feature maps. Please refer to the original paper \cite{CBAM} for complete and detailed descriptions of SAM and CAM. To further clarify the structure composition of ParaNet-SC, we list all the included major components as well as the corresponding output dimensions in Table \ref{Sup-Table-1}. 

\subsection*{2.3 ParaNet-PU: Downstream Task Network for Point Cloud Upsampling}
The ParaNet-PU framework is designed to achieve $4 \times$ point cloud upsampling following a patch-based processing pipeline. Specifically, we tend to upsample a $256$-point sparse patch to a $1024$-point dense patch through ParaNet-PU. As described in the paper, we explored three different types of training strategies (\textit{i.e.}, JO, Off-IO, and On-IO) for connecting ParaNet with a specific downstream task network. For the shape classification task, we can observe that the JO scheme outperforms the other two choices. Although joint optimization always results in higher information loss, \textit{i.e.}, more points may be lost in the generated PGI, it benefits from end-to-end training in which the PGI representation can be adaptively adjusted to some extent, boosting the overall classification performance. Different from shape classification that is relatively less sensitive to information loss, point cloud upsampling highly relies on the original geometry and topology information provided by the input sparse point set. Even if only a few critical points are missing, it would be rather difficult to accurately infer the underlying surface. Thus, we adopted the On-IO scheme for training our ParaNet-PU framework, generating PGIs with the resolution of $128 \times 128$ for all training and testing patches with $256$ points.

Technically, we started by feeding the input PGI into a series of convolutional layers with large sized filters followed by max-pooling operations to reduce feature resolution. We introduced self-attention, as proposed in \cite{wang2018non}, to build long-range point dependency and encode the global shape prior of the input patch. After that, we employed the back projection mechanism \cite{BackProjection} to fully exploit the mutual relationships of low- and high-resolution feature maps. As investigated in \cite{BackProjection}, the back projection workflow involves the up-projection units (UPU) and the down-projection units (DPU). Following the original design, we sequentially organized the two types of projection units, alternating between the UPU and the DPU, which can be considered as a self-correcting learning procedure providing a beneficial error feedback mechanism. In Table \ref{Sup-Table-2}, we summarize the major components and the corresponding output dimensions. As suggested, we concatenated the up-sampled feature maps produced by the three UPUs and then applied $1 \times 1$ convolutions to generate 3D Cartesian coordinates. In the end, we could easily reshape the predictions into a 2D array, representing the upsampled point cloud patch.

During training, we employed the Earth Mover's Distance (EMD) to measure the difference between upsampling results and ground truths. To promote the uniformity of the generated patches, we additionally imposed a point distribution constraint, expecting that the nearest neighbors are neither too close nor too far from each other. To achieve this, we adopted a modified \textit{repulsion loss}, which was originally proposed in PU-Net \cite{PU-Net}. Given an upsampled patch $\mathbf{X}=\{ {x_i} \}_{i=1}^{U}$ and the corresponding ground truth patch $\mathbf{Y}=\{ {y_i} \}_{i=1}^{U}$, we formulated the uniformity constraint in a fast-decaying manner as follows

\begin{equation}
	\mathcal{L}_{uni} = \gamma \cdot \frac{1}{U} \sum_{i=1}^{U} -r(x_i) \cdot \exp(-\frac{{r(x_i)}^2}{2h^2}),
\end{equation}
where $r(x_i)$ represents the Euclidean distance between $x_i$ and its nearest neighbor, and $\gamma$ controls the contribution of $\mathcal{L}_{uni}$ when combined with the EMD loss function for point generation supervision. Empirically, we set the parameter $\gamma$ to be $0.01$. Actually, the above constraint tends to push the nearest neighbor distance $r(x_i)$ to a fixed value of $h$, which is chosen as the average nearest neighbor distance in patch $\mathbf{Y}$ as follows
\begin{equation}
	h= \frac{1}{U} \sum_{i=1}^{U} r(y_i).
\end{equation}

To further boost the generation of uniformly distributed point clouds, we introduced side supervisions to enhance the learned intermediate feature representations. In our implementation, we applied additional $1 \times 1$ convolutions to the outputs of DPU-1 and DPU-2, predicting sparse patches with $256$ points. Then we used farthest point sampling to downsample the corresponding ground truth patch with $1024$ points into a uniform $256$-point patch, which serves as the intermediate supervision information to guide the predicted side outputs using the weighted combination of EMD and uniformity constraint.

\begin{table}[!t]
	\renewcommand\arraystretch{1.1}
	\caption{Components and outputs of ParaNet-PU.}
	\begin{center}
		\setlength{\tabcolsep}{3mm}{
			\begin{tabular}{ c | c }
				\hline\hline
				Components & Outputs ([Channels, Height, Width]) \\
				\hline
				PGI    & [3, 128, 128] \\
				Conv ($12 \times 12$) & [64, 64, 64]  \\
				Conv ($8 \times 8$) & [128, 32, 32]  \\
				Self-Att & [128, 32, 32]  \\
				Conv ($6 \times 6$) & [256, 16, 16]  \\
				Self-Att & [256, 16, 16]  \\
				UPU-1 & [256, 32, 32] \\
				DPU-1 & [256, 16, 16] \\
				UPU-2 & [256, 32, 32] \\
				DPU-2 & [256, 16, 16] \\
				UPU-3 & [256, 32, 32] \\
				Up-Concat & [768, 32, 32] \\
				Conv ($1 \times 1$) & [3, 32, 32] \\
				Reshape & [3, 1024] \\
				\hline\hline
		\end{tabular}}
	\end{center}
	\label{Sup-Table-2}
\end{table}

\section*{3. Backbone Point Feature Extractor}

\begin{table}[!t]
	\renewcommand\arraystretch{1.1}
	\caption{Comparison of Different Point Feature Extractors for 3D Shape Classification.}
	\begin{center}
		\setlength{\tabcolsep}{5mm}{
			\begin{tabular}{ c | c | c }
				\hline\hline
				Method & Rep Acc (CD) & Cls Acc (\%) \\
				\hline
				ParaNet-SC-DGCNN (JO) & $2.32 \times 10^{-6}$ & 93.1 \\
				ParaNet-SC-PointNet (JO) & $9.61 \times 10^{-6}$ & 93.1 \\
				\hline\hline
		\end{tabular}}
	\end{center}
	\label{Sup-Table-3}
\end{table}

\begin{table}[!t]
	\renewcommand\arraystretch{1.1}
	\caption{Comparison of Different Point Feature Extractors for Point Cloud Upsampling.}
	\begin{center}
		\setlength{\tabcolsep}{3.0mm}{
			\begin{tabular}{ c | c | c | c | c}
				\hline\hline
				Method & Rep Acc (CD) & P2F & CD & HD \\
				& $(10^{-8})$ & $(10^{-3})$ & $(10^{-3})$ & $(10^{-3})$ \\
				\hline
				ParaNet-PU-DGCNN (On-IO) & 2.09 & 2.26 & 0.28 & 4.44  \\
				ParaNet-PU-PointNet (On-IO) & 7.15 & 2.27 & 0.28 & 4.49  \\
				\hline\hline
		\end{tabular}}
	\end{center}
	\label{Sup-Table-4}
\end{table}

In our paper, we employed DGCNN \cite{DGCNN} as the backbone point feature extractor of ParaNet. Actually, we argue that DGCNN can be seamlessly replaced by any other kind of deep set architecture, and the overall performance over a specific downstream application is dominated by the task network and less sensitive to the point feature extractor. To empirically demonstrate this point, we replaced DGCNN \cite{DGCNN} with PointNet \cite{PointNet} in the ParaNet for generating PGIs while maintaining the subsequent components. In Tables \ref{Sup-Table-3} and \ref{Sup-Table-4}, we compared the data representation accuracy (Rep Acc) under the Chamfer distance (CD) metric and the overall task performance of DGCNN-driven (\textit{i.e.}, ParaNet-SC/PU-DGCNN) and PointNet-driven (\textit{i.e.}, ParaNet-SC/PU-PointNet) processing pipelines. We can observe that although PointNet, a backbone point encoder which is not as powerful as DGCNN, results in some decrease in the data representation accuracy, the overall performances of the downstream tasks are basically not affected, which implies that the proposed PGI-driven point cloud processing framework is not limited by a specific type of point feature extractor. In practice, we can expect to focus on exploiting and designing generic CNN architectures to promote various point cloud applications.
\end{document}